\documentclass[11pt]{article}
\usepackage{amssymb}
\usepackage{graphicx}
\usepackage{amsfonts}
\usepackage{amsmath}
\usepackage{longtable,lscape}
\usepackage{amscd}
\usepackage{mathrsfs}
\usepackage{latexsym}
\usepackage{bbm}
\usepackage{float}
\usepackage{url}

\newcommand{\captionfonts}{\footnotesize}
\makeatletter  
\long\def\@makecaption#1#2{%
  \vskip\abovecaptionskip
  \sbox\@tempboxa{{\captionfonts #1: #2}}%
  \ifdim \wd\@tempboxa >\hsize
    {\captionfonts #1: #2\par}
  \else
    \hbox to\hsize{\hfil\box\@tempboxa\hfil}%
  \fi
  \vskip\belowcaptionskip}
\makeatother 

\title{{\bf Quantum Structures in Human Decision-making: Towards Quantum Expected Utility}}
\author{Sandro Sozzo\\ School of Business and Centre IQSCS \\ University Road LE1 7RH \\ 
Leicester (United Kingdom) \\ Email address: \url{ss831@le.ac.uk}}

\date{}

\begin{document}

\maketitle

\begin{abstract}
\noindent
{\it Ellsberg thought experiments} and empirical confirmation of Ellsberg preferences pose serious challenges to {\it subjective expected utility theory} (SEUT). We have recently elaborated a quantum-theoretic framework for human decisions under uncertainty which satisfactorily copes with the Ellsberg paradox and other puzzles of SEUT. We apply here the quantum-theoretic framework to the {\it Ellsberg two-urn example}, showing that the paradox can be explained by assuming a state change of the conceptual entity that is the object of the decision ({\it decision-making}, or {\it DM}, {\it entity}) and representing subjective probabilities by quantum probabilities. We also model the empirical data we collected in a DM test on human participants within the theoretic framework above. The obtained results are relevant, as they provide a line to model real life, e.g., financial and medical, decisions that show the same empirical patterns as the two-urn experiment. 
\end{abstract}

\section{Introduction}\label{intro}
Daniel Kahneman was awarded the Nobel Prize in Economic Science in 2002 for his pioneering studies on the identification and estimation of the psychological factors that influence human behaviour under uncertainty, which led to the birth of a new domain called {\it behavioural economics}.

Cognitive psychologists have assumed for years, often implicitly, that complex cognitive processes, like human judgement and decision-making (DM), have to be modelled by combining set-theoretic structures and should obey to mathematical relations that resemble those typically used in logic, formalized by Boole ({\it Boolean logic}), and probability, axiomatized by Kolmogorov ({\it Kolmogorovian probability}) \cite{k1933}. These structures are known in physics as {\it classical structures}: they were originally used in classical physics, and later extended to statistics, psychology, economics, finance and computer science. Classical structures are also implicitly assumed in the so-called {\it Bayesian approach}, according to which any source of uncertainty can be formalized probabilistically, while people update knowledge according to the {\it Bayes law} of Kolmogorovian probability. Finally, classical structures are the building blocks of {\it subjective expected utility theory} (SEUT), providing both the descriptive and the normative foundations of rational DM: in situations of uncertainty, people (should) choose as if they maximized EU with respect to a unique probability measure, satisfying the axioms of Kolmogorov and interpreted as their {\it subjective probability} \cite{s1954,gps2008}.

However, on the one side, empirical research in cognitive psychology has revealed that classical structures are not generally able to model human judgements and decisions, thus making problematical the interpretation of a wide range of cognitive phenomena in terms of standard logic and probability theory. On the other side, Kahneman, Tversky and other authors suggested that these empirical deviations from classicality are ``true errors'' of human reasoning, whence the use of terms like ``effect'', ``fallacy'', ``paradox'', ``contradiction'', etc., to refer to such phenomena \cite{kst1982,kt2000}.\footnote{Empirical psychology usually distinguishes between {\it probability judgement errors} and {\it DM errors}. Conjunctive and disjunctive fallacies in Linda-like stories, over- and under-extension effects in membership judgements on conceptual combinations are examples of the former category. The disjunction effect, the prisoner's dilemma and the paradoxes of SEUT studied here are instead examples of the latter category \cite{bb2012}.}

An innovative aspect of the Kahneman-Tversky programme of judgement heuristics and individual biases was the use of non-Kolmogorovian structures to represent probabilities. In that regard, the research on the foundations of quantum theory has unveiled both the conceptual and mathematical differences between classical and quantum structures, e.g., context-dependent situations crucially need a non-Kolmogorovian quantum-like model of probability (see, e.g., \cite{a2009,abgs2013,ags2013}). This was the starting point of a successful research that applies the mathematical formalism of quantum theory, detached from any physical interpretation, to model situations in cognition and economics that cannot be modelled by classical structures (see, e.g., \cite{bb2012,a2009,abgs2013,ags2013,hk2013}).

Why are classical structures problematical in SEUT? In 1961, Daniel Ellsberg proposed a series of DM experiments whose results do not agree with the predictions of SEUT, as far as concrete decisions are mainly influenced by psychological factors, like {\it ambiguity aversion}, rather than by the need of maximizing EU \cite{e1961}. The consequence is that one cannot represent Ellsberg preferences by maximization of EU with respect to a unique Kolmogorovian probability measure, which generates the famous {\it Ellsberg paradox}. Ellsberg preferences have been empirically confirmed several times against the predictions of SEUT and its main extensions, in simple DM tests, but also in more complex real life DM situations (see, e.g., \cite{ms2014}).  

How does the research on quantum structures relate to the pitfalls of SEUT? The answer comes from a 20-year research on the foundations of quantum physics, the origin of quantum structures and the conceptual and mathematical differences between classical and quantum theories. Indeed, we have followed the operational and realistic approach to quantum physics elaborated by the Brussels research team, extending it to cognitive entities, that is, concepts, combinations of concepts, propositions and more complex DM entities \cite{asdbs2016}. In this approach, the situation that is the object of the decision defines a {\it conceptual entity} (the {\it DM entity}) in a given {\it state}. This state has a cognitive nature and captures conceptual and mathematical aspects of the uncertainty (risk, ambiguity) surrounding the decision situation. When the decision-maker is asked to choose between different options, her/his mental action is described as a cognitive {\it context} influencing the DM entity and changing its state. The way in which the state of the DM entity changes due to this  contextual interaction provides information about the decision-maker's attitude towards ambiguity (ambiguity attraction, ambiguity seeking, ambiguity neutral).

The above realistic and operational description of the DM process as a ``constructive (creation) process in which a context-mediated transition occurs from potential to actual'' closely resembles a quantum measurement process in which the interaction with the physical measurement context determines an intrinsically probabilistic state change of the measured quantum entity, which suggests that a quantum mechanical representation is appropriate in this case. In this regard, we have elaborated a quantum-theoretic framework to represent human DM under uncertainty that enables modelling of the Ellsberg paradox \cite{ast2014,as2016,ahs2017} and more general DM situations \cite{ast2014,ahs2017} that are problematical from the point of view of SEUT and its extensions, i.e. the {\it Machina paradox} \cite{m2009}. Moreover, we have recently proved that the quantum-theoretic framework enables successful representation of various DM tests performed on the Ellsbeg and Machina paradox situations \cite{agms2018}.

In the present paper, we focus on the quantum-theoretic modelling of the {\it Ellsberg two-urn experiment}, inasmuch this experiment involves a choice between an option with a known probability, or {\it risky option}, and a choice with an unknown probability, or {\it ambiguous option}. Moreover, the design of this experiment is very similar to more general tests, involving financial and medical decisions, in which a shift is observed from the risky option to the ambiguous one, and viceversa, depending on the amount of uncertainty surrounding the decision. The quantum-theoretic framework predicts the possibility of such a shift, hence it is able to model more complex real life situations, as we demonstrate in Sect. \ref{conclusions}. 

For the sake of completeness, let us summarize the content of this paper, as follows.

In Sect. \ref{SEUT}, we give an overview of the main notions and results of SEUT, also pointing out its main empirical  pitfalls. In particular, we present a DM test of the Ellsberg two-urn example we performed, which reveals a strong ambiguity aversion pattern. In Sect. \ref{QEUT}, we sketch the essential results of our quantum-theoretic framework. In particular, we stress that the quantum-theoretic approach suggests the development of a state-dependent quantum-based SEUT in which decision-makers maximize EU with respect to a quantum probability measure. In Sect. \ref{quantumellsberg}, we apply the quantum-theoretic framework to the two-urn example, reproducing Ellsberg preferences and faithfully representing the experimental data of Sect. \ref{SEUT}. Finally, in Sect. \ref{conclusions}, we consider some DM tests on managerial and medical decisions \cite{vc1999,hkk2002}, which can be formalized exactly as the two-urn example, and we demonstrate that its results can be naturally modelled within the quantum model for the two-urn experiment.

\section{Expected utility theory and its descriptive pitfalls}\label{SEUT}
The first axiomatic formulation of EUT can be traced back to the seminal work of von Neumann and Morgenstern: in the presence of uncertainty, people choose in such a way to maximize their EU with respect to a unique Kolmogorovian probability measure \cite{vnm1944}. However, this formulation of EUT only deals with the uncertainty that can be described by known probabilities ({\it objective uncertainty}, or {\it risk}). On the other hand, situations frequently occur in which uncertainty cannot be described by known probabilities ({\it subjective uncertainty}, or {\it ambiguity}) \cite{k1921}. The Bayesian approach mentioned in Sect. \ref{intro} simplifies the distinction between objective and subjective uncertainty introducing the notion of {\it subjective probability}: even when probabilities are not known, people may still make their own {\it beliefs}, or {\it priors}, which may differ from one person to another, and with respect to which they maximize EU. As a matter of fact, Savage presented an axiomatic formulation of SEUT which extends the one of von Neumann and Morgenstern and perfectly accords with the Bayesian approach \cite{s1954}.

We introduce here the basic definitions and results of SEUT we need to obtain our results in Sects. \ref{QEUT} and \ref{quantumellsberg}. The reader who is interested to deepen these notions can refer to \cite{s1954,ms2014,gm2013}. Let:

(i) ${\mathscr S}$ be the set of all (physical) {\it states of nature}, assumed discrete and finite, for the sake of simplicity;

(ii) ${\mathscr P}({\mathscr S})$ be the power set of ${\mathscr S}$;

(iii) ${\mathscr A}\subseteq {\mathscr P}({\mathscr S})$ be a $\sigma$-algebra, whose elements denote {\it events};

(iv) $p:{\mathscr A}\subseteq {\mathscr P}({\mathscr S})\longrightarrow [0,1]$ be a Kolmogorovian probability measure over $\mathscr A$;

(v) ${\mathscr X}$ be the set of all consequences, whose elements are assumed to denote {\it monetary payoffs}, for the sake of simplicity;

(vi) $f: {\mathscr S} \longrightarrow {\mathscr X}$ be a function denoting an {\it act};

(vii) $\mathscr{F}$ be the set of all acts;

(viii) $\succsim$ be a weak preference relation, i.e. reflexive, symmetric and transitive over the Cartesian product $\mathscr{F} \times \mathscr{F}$, where $\succ$ and $\sim$ are the strong preference and indifference relations, respectively;\footnote{For example, assuming an approach in which preferences are {\it revealed}, if a person strongly, or strictly, prefers act $f$ to act $g$, we write $f \succ g$. Analogously, if a person is indifferent between $f$ and $g$, we write $f \sim g$.}

(ix) $u: {\mathscr X} \longrightarrow \Re$ be an {\it utility function}, assumed strictly increasing and continuous, as in the traditional literature;

(x) $\{E_1, \ldots, E_n \}$ be a set of mutually exclusive and exhaustive elementary events forming a partition of ${\mathscr S}$;

(xi) for every $i \in \{ 1, \ldots, n\}$, $x_i$ be the utility associated by the act $f$ to the event $E_i$, i.e. $f=(E_1,x_1;\ldots;E_n,x_n)$;

(xii) $W(f)=\sum_{i=1}^{n}p(E_i)u(x_i)$ be the {\it expected utility} associated with the act $f$ with respect to the Kolmogorovian probability measure $p$. 

Savage proved that, if the algebraic structure $({\mathscr F}, \succsim)$ satisfies suitable axioms\footnote{One of the axioms is the famous {\it sure thing principle}, which is violated in the Ellsberg paradox. Other axioms have a technical nature. However, the axioms of SEUT are not relevant to the present purposes, hence we will not dwell on them, for the sake of brevity.} then, for every $f,g\in \mathscr F$, a unique Kolmogorovian probability measure $p$ and a unique (up to positive affine transformations) utility function $u$ exist such that $f$ is preferred to $g$, i.e. $f \succsim g$, if and only if the EU of $f$ is greater than the EU of $g$, i.e. $W(f) \ge W(g)$. For every $i\in \{ 1, \ldots, n\}$, the utility value $u(x_i)$ depends on the decision-maker's risk preferences, while $p(E_i)$ is interpreted as the subjective probability that the event $E_i$ occurs \cite{s1954}.

Savage's result above is testable at a descriptive level, in the sense that it can be empirically tested through human decisions, and compelling at a normative level, in the sense that it prescribes what ``rational behaviour'' should be. However, Daniel Ellsberg proposed in 1961 a series of thought experiments in which, he suggested, decision-makers are not likely to maximize EU, rather they are likely to prefer acts with known (or objective) probabilities over acts with unknown (or subjective) probabilities. While a famous thought experiment is the {\it three-color example}, we prefer here to discuss the {\it two-urn example}, for reasons that will become clear in Sect. \ref{conclusions}. 

Consider two urns, urn I with 100 balls that are either red or black in unknown proportion, and urn II exactly with 50 red balls and 50 black balls. One ball is to be drawn at random from each urn. Then, free of charge, a person is asked to bet on pairs of the acts $f_1$, $f_2$, $f_3$ and $f_4$ in Table 1.

\noindent 
\begin{table} \label{table02}
\begin{center}
\begin{tabular}{|p{1.5cm}|p{1.5cm}|p{1.5cm}||p{1.5cm}|p{1.5cm}|}
\hline
\multicolumn{1}{|c|}{} & \multicolumn{2}{c||}{Urn I} & \multicolumn{2}{c|}{Urn II} \\
\hline
\multicolumn{1}{|c|}{} & \multicolumn{2}{c||}{}  & \multicolumn{1}{c|}{1/2} & \multicolumn{1}{c|}{1/2}  \\
\hline
Act & Red & Black & Red & Black \\ 
\hline
\hline
$f_1$ & \$100 & \$0 &  &  \\ 
\hline
$f_2$ &  &  & \$100 & \$0 \\ 
\hline
$f_3$ & \$0 & \$100 &  &  \\ 
\hline
$f_4$ &  &  & \$0 & \$100 \\ 
\hline
\end{tabular}
\end{center}
{\bf Table 2.} The payoff matrix for the Ellsberg two-urn example.
\end{table}
Ellsberg suggested that most people will generally prefer  $f_2$ over $f_1$ and $f_4$ over $f_3$. The reason for this choice is simple: $f_2$ and $f_4$ are unambiguous acts, because they are associated with events over known probabilities, 0.5 in this case, while $f_1$ and $f_3$ are ambiguous acts, because they are associated with events over unknown probabilities, ranging from 0 to 1 in this case \cite{e1961}. This attitude of decision-makers towards ambiguity is called {\it ambiguity aversion}. Several experiments on the two-urn example have confirmed the {\it Ellsberg preferences} $f_2 \succ f_1$ and $f_4 \succ f_3$, hence an ambiguity aversion attitude of decision-makers (see, e.g., \cite{ms2014}).

The behaviour of a decision-maker who is psychologically influenced by ambiguity cannot be reproduced by SEUT. Indeed, assuming that decision-makers assign subjective probabilities $p_R$ and $p_B=1-p_R$ to the events ``a ball of red color is drawn'' and ``a ball of color black is drawn'', respectively, then the condition $W(f_2)>W(f_1)$ is equivalent to $(p_R-\frac{1}{2})(u(100)-u(0))<0$, where $u(0)$ and $u(100)$ denote the utilities associated with the payoffs 0 and 100, respectively. On the contrary, the condition $W(f_4)>W(f_3)$ is equivalent to $(p_R-\frac{1}{2})(u(100)-u(0))>0$. Hence, one cannot find a Kolmogorovian probability measure such that $f_2 \succ f_1$ and $f_4 \succ f_3$ by maximization of the EU functional with respect to that measure, whence the paradox.

The Ellsberg paradox above and other Ellsberg-type puzzles put at stake both the descriptive and the normative foundations of SEUT, which led various scholars to propose alternatives to SEUT, in which more general, and even non-Kolmogorovian, mathematical structures are used to represent subjective probabilities. Major non-EU models include {\it rank dependent EU}, {\it EU with multiple priors}, {\it second order beliefs}, etc. (see, e.g., \cite{ms2014,gm2013}). 

In 2009, Mark Machina elaborated two variants of the Ellsberg paradox, the {\it 50/51 example} and the {\it reflection example}, which challenge major non-EU models in a similar way as the Ellsberg examples challenge SEUT \cite{m2009,blhp2011}. {\it Machina preferences} have been confirmed in two tests against the predictions of both SEUT and its non-EU generalizations \cite{agms2018} and \cite{lhp2010}. The implication of Ellsberg and Machina paradoxes is that a unified theoretic approach to represent human preferences and choices under uncertainty is still an unachieved goal \cite{lhp2010}.

We illustrate and analyse here a DM test we performed on the Ellsberg two-urn example which confirms traditional ambiguity aversion patterns and enables a quantum-theoretic modelling \cite{agms2018}. The results and ensuing quantum model have far reaching implications on applications of ambiguity aversion to financial and medical decisions, as we will see in Sect. \ref{conclusions}.

We presented a sample of 200 people with a questionnaire in which they had to choose between the pairs of acts ``$f_1$ versus $f_2$'' and ``$f_3$ versus $f_4$'' in Table 1. Respondents had overall a basic knowledge of probability theory, but no specific training in decision theory. Respondents were provided with a paper similar to the one in Figure 1.\footnote{For the sake of simplicity, we assumed that each choice concerned two alternatives, hence indifference between acts was not a possible option.}
\begin{figure}
\begin{center}
\includegraphics[scale=0.5]{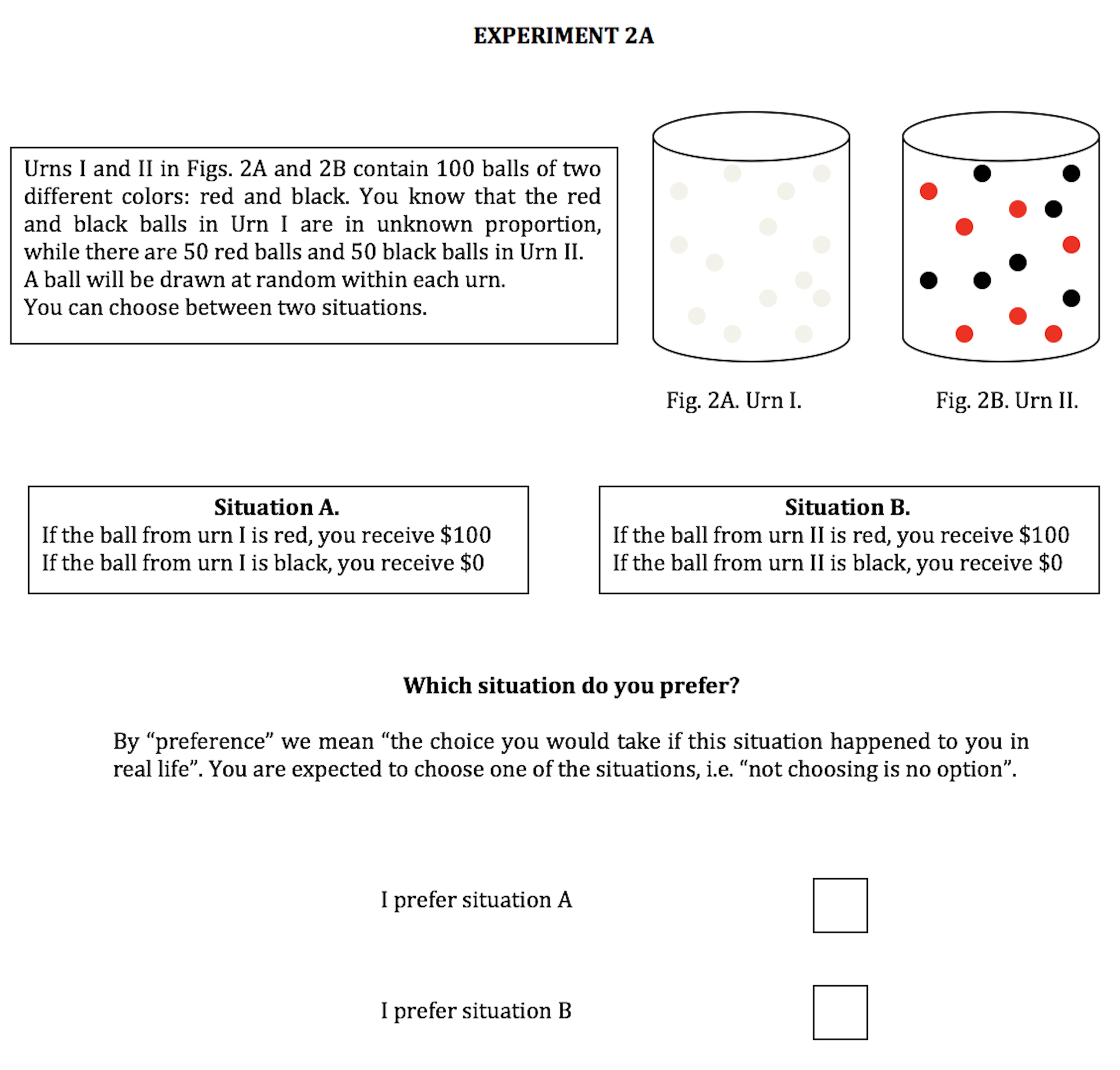}
\end{center}
{\bf Figure 1.} A sample of the questionnaire on the Ellsberg two-urn experiment. It corresponds to the choice between acts $f_1$ and $f_2$ in Table 1.
\end{figure}
In the test, 26 respondents chose acts $f_1$ and $f_3$, 10 chose $f_1$ and $f_4$, 6 chose $f_2$ and $f_3$, and 158 chose $f_2$ and $f_4$. Equivalently, 164 respondents over 200 preferred act $f_2$ over act $f_1$, for a preference rate of 164/200=0.82 (the difference is significant, $p=1.49E-24$). Moreover, 168 respondents over 200 preferred act $f_4$ over act $f_3$, for a preference rate of 168/200=0.84 (the difference is again significant, $p=1.25E-28$). Finally, 184 respondents over 200 preferred either $f_1$ and $f_4$ or $f_2$ and $f_3$, for an inversion rate of 184/200=0.92. This pattern agrees with the Ellsberg preferences and points towards ambiguity aversion, while it cannot be reproduced by SEUT. In addition, these results confirm empirical findings existing in the literature (see, e.g., \cite{ms2014}).

It is important to notice that in each pair of acts, decision-makers are asked to choose between a risky option, that is, an option with known probability of getting a given consequence, and an ambiguous option, that is, an option with unknown probability of getting the same consequence. We will see in Sect. \ref{conclusions} that this is exactly the experimental setting that is designed to test individual attitudes towards ambiguity in more concrete DM situations, like medical and financial decisions. This makes relevant a systematic study of the two-urn example and the ensuing modelling within the quantum mathematical formalism.

We sketch in the next section the quantum-theoretic framework which models human decisions, ambiguity, ambiguity aversion, and explains Ellsberg and other puzzles of SEUT in terms of genuine quantum structures.

\section{A quantum-theoretic framework for expected utility}\label{QEUT}
The quantum-theoretic framework we summarize in this section provides a successful modelling of the Ellsberg and Machina paradox situations \cite{as2016}. In addition, it enables quantum representation of various sets of DM tests on these paradoxes, namely, the three-color experiment \cite{ast2014,agms2018}, the 50/51 experiment \cite{ast2014,agms2018} and the reflection experiment \cite{ahs2017,agms2018}. The quantum-theoretic framework constitutes the first step towards the development of a quantum-based state-dependent extension of SEUT \cite{ahs2017}.

Let us start by introducing the basic notions we need for attaining our purposes. These notions rest on the realistic and operational approach worked out by the Brussels research team for both quantum and cognitive entities \cite{asdbs2016}.

The object of the decision, the decision-maker interacts with, defines a {\it conceptual DM entity} which is supposed to be in a defined state $p_v$. Such a state captures aspects of ambiguity and has a conceptual nature, thus it must be distinguished from a physical state of nature (see Sect. \ref{SEUT}). Let $\Sigma_{DM}$ denote the set of all states of the DM entity. Let ${\mathscr E}$ denote the set of all events which may occur when the DM entity is in a specific state. For a state $p_v\in \Sigma_{DM}$, let $\mu(E,p_v)$ denote the (subjective) probability that $E$ occurs when the DM entity is in the state $p_v$.

Then, let $E_1, E_2, \ldots, E_n \in \mathscr E$ denote mutually exclusive and exhaustive elementary events, let ${\mathscr X}\subseteq \Re$ be the set of consequences (monetary outcomes), and let, for every $i \in \{ 1,\dots,n\}$, the act $f$ map the elementary event $E_i \in \mathscr E$ into the outcome  $x_i\in \Re$, so that $f=(E_1,x_1;\ldots;E_n,x_n)$.\footnote{The intepretation of the symbol $f=(E_1,x_1;\ldots;E_n,x_n)$ is the usual one in decision theory, namely, we get $x_1$ if the event $E_1$ occurs, \ldots, $x_n$ if the event $E_n$ occurs.} Finally, let $u: {\mathscr X} \longrightarrow \Re$ be a continuous strictly increasing utility function expressing individual preferences towards risk.

Like in the canonical representation of quantum entities, we associate the DM entity with a Hilbert space $\mathscr H$ over the field $\mathbb C$ of complex numbers. The number $n$ of mutually exclusive and exhaustive elementary events entails that the Hilbert  space $\mathscr H$ can be chosen to be isomorphic to the Hilbert space ${\mathbb C}^n$ of all n-tuples of complex numbers. Let $\{ |\alpha_1\rangle, |\alpha_2\rangle, \ldots, |\alpha_n\rangle\}$ be the canonical orthonormal (ON) basis of ${\mathbb C}^n$, i.e. $|\alpha_1\rangle=(1,0,\ldots 0)$, \ldots, $|\alpha_n\rangle=(0,0,\ldots n)$. 

The set $\mathscr E$ of events is represented by the complete orthocomplemented, but non-distributive, lattice ${\mathscr L}( {\mathbb C}^n)$  of all orthogonal projection operators over ${\mathbb C}^n$. In particular, for every $i \in \{1,\ldots, n \}$, the elementary event $E_i$ is represented by the 1-dimensional orthogonal projection operator $P_i=|\alpha_i \rangle \langle \alpha_i|$.

For every state $p_v \in \Sigma_{DM}$ of the DM entity, represented by the unit vector $|v\rangle=\sum_{i=1}^{n}\langle \alpha_i|v\rangle |\alpha_i\rangle\in {\mathbb C}^n$, the mapping
\begin{equation}
\mu_{v}: P \in {\mathscr L}( {\mathbb C}^n) \longmapsto  \mu_{v}(P)\in [0,1]
\end{equation}
induced by the Born rule is a quantum probability measure over ${\mathscr L}( {\mathbb C}^n)$. In particular, $\mu_{v}(P)$ is identified with the (subjective) probability $\mu(E,p_v)$ that the event $E$, represented by the orthogonal projection operator $P$, occurs when the DM entity is in the state $p_v$. Thus, in particular, for every $i\in \{ 1, \ldots, n\}$,
\begin{equation} \label{quantumprobability}
\mu(E_i,p_v)=\langle v|P_i|v\rangle=|\langle \alpha_i|v\rangle|^{2}
\end{equation}

Let us now represent acts by using the quantum mechanical formalism. The act $f=(E_1,x_1;\ldots;E_n,x_n)$ is represented by the Hermitian operator
\begin{equation} \label{quantumact}
\hat{F}=\sum_{i=1}^{n} u(x_i)P_i=\sum_{i=1}^{n} u(x_i)|\alpha_i\rangle\langle \alpha_i|
\end{equation}
For every $p_v \in \Sigma_{DM}$, we introduce the functional ``EU in the state $p_v$'', $W_v:{\mathscr F} \longrightarrow \Re$, as follows. For every $f \in {\mathscr F}$,
\begin{eqnarray} \label{quantumexpectedutility}
W_{v}(f)&=&\langle v| \hat{F}| v \rangle=\langle v| \Big ( \sum_{i=1}^{n} u(x_i)P_i   \Big ) |v\rangle \nonumber \\
=\sum_{i=1}^{n} u(x_i) \langle v|P_i|v\rangle&=&\sum_{i=1}^{n} u(x_i) |\langle \alpha_i|v\rangle|^{2}=\sum_{i=1}^{n} \mu(E_i,p_v)u(x_i)
\end{eqnarray}
where we have used (\ref{quantumprobability}) and (\ref{quantumact}). Equation (\ref{quantumexpectedutility}) generalizes the SEUT formula in (xii), Sect. \ref{SEUT}. We note that the EU generally depends on the state $p_v$ of the DM entity. When $W_v(f)$ does (not) explicitly depend on the state $p_v$ of the DM entity, then the act $f$ is (un)ambiguous. Thus, $p_v$ mathematically and conceptually incorporates the presence of ambiguity. This means in particular that, for every $f,g\in \mathscr F$, states $p_v,p_w\in \Sigma_{DM}$ may exist such that $W_{v}(f)>W_{v}(g)$,  whereas $W_{w}(f)<W_{w}(g)$, depending on decision-makers' attitudes towards ambiguity. This suggests introducing a state-dependent preference relation $\succsim_{v}$ on the set of acts $\mathscr F$, as follows. 

For every $f,g \in {\mathscr F}$ and $p_{v}\in \Sigma_{DM}$,
\begin{equation} \label{statedep}
f \succsim_{v} g \ {\rm iff} \ W_{v}(f) \ge W_{v}(g)
\end{equation}
Equation (\ref{statedep}) indicates that one could in principle identify a set of axioms on acts and their order relations allowing to uniquely represent preferences by maximization of a state-dependent EU functional, thus opening the way towards a quantum SEUT. However, such a {\it representation theorem} would go beyond the scopes of the present paper, hence we do not discuss it here.

Let us come to the DM process. The state of the DM entity can change under the effect of a context, which has again a cognitive nature. An example of such a cognitive context is a measurement with $n$ possible outcomes $\{ 1, \ldots, n\}$ that can be performed on the DM entity, where the $i$-th outcome is associated with the elementary event $E_i \in \mathscr E$. If the $i$-th outcome is obtained, the state of the DM entity is transformed  into a state represented by the unit vector $|\alpha_i\rangle$. More generally, if we denote the set of all contexts by $\mathscr C$, then their influence on the DM entity can be described by the transition probability $\nu(p_v,c,p_{v_0})$, where the context $c \in \mathscr C$ changes the initial state $p_{v_0}\in \Sigma_{DM}$ into the final state $p_v\in \Sigma_{DM}$.

Suppose now that, when the decision-maker is presented with a questionnaire involving a choice between the acts $f$ and $g$, the DM entity is in the initial state $p_{v_0}$, which is generally determined by symmetry reasons. The state $p_{v_0}$ is interpreted as the state of the DM entity in the absence of any context (equivalently, in the presence of the unitary context). As the decision-maker starts comparing $f$ and $g$, this cognitive action can be described as a context interacting with the DM entity and changing its state. The type of state change directly depends on the decision-maker's attitude towards ambiguity. More precisely, a given attitude towards ambiguity, say ambiguity aversion, will determine a given change of state of the DM entity to a state $p_v$, inducing the decision-maker to prefer, say $f$. But, a different attitude towards ambiguity, say ambiguity seeking, will determine a different change of state of the DM entity to a state $p_{w}$, leading the decision-maker to instead prefer $g$. In this way, different attitudes towards ambiguity are formalized by different changes of state of the DM entity hence, through (\ref{quantumprobability}), by different subjective probability measures.

We proved in \cite{ahs2017,agms2018} that the state-dependence above can exactly explain the {\it inversion of preferences} observed in the Ellsberg and Machina paradox situations. The conclusion is interesting, from our point of view. On the one side, SEUT claims that decision makers should choose in such a way to maximize EU with respect to a Kolmogorovian probability measure. On the other side, we found that various puzzles of SEUT are solved if one assumes that decision makers actually maximize EU with respect to a non-Kolmogorovian, specifically quantum, probability measure.

\section{An application to the Ellsberg two-urn example}\label{quantumellsberg}
In this section, we specify the quantum-theoretic framework of Sect. \ref{QEUT} to the Ellsberg two-urn example, showing that it enables faithful representation of the experimental data in Sect. \ref{SEUT}.

The two-urn example defines two conceptual entities, {\it DM entity I}, which is the urn with 100 red or black balls in unknown proportion, and {\it DM entity II}, which is the urn with 50 red balls and 50 black balls. In the absence of any context, for uniformity reasons, we can assume that both entities are initially in the state $p_{v_0}$, which has a cognitive nature, as we have seen in Sect. \ref{QEUT}. 

Let $E_R$ and $E_B$ denote the exhaustive and mutually exclusive elementary events ``a red ball is drawn'' and ``a black ball is drawn'', respectively. They define a ``color measurement context'' on both DM entity I and DM entity II with two outcomes, corresponding to the colors red and black. 

Thus, both DM entity I and DM entity II are associated with the 2-dimensional complex Hilbert space ${\mathbb C}^{2}$. Let $\{(1,0), (0,1) \}$ be the canonical ON basis of ${\mathbb C}^{2}$. The color measurement is represented by a Hermitian operator with eigenvectors $|R\rangle=(1,0)$ and $|B\rangle=(0,1)$ or, equivalently, by the spectral family $\{ P_{R}=|R\rangle\langle R|, P_{B}=|B\rangle\langle B|=\mathbbmss{1}-P_R\}$.  In the canonical basis of ${\mathbb C}^{2}$, the initial state $p_{v_0}$ of both DM entity I and DM entity II is represented, due to the uniformity reasons above, by the unit vector
\begin{equation}
|v_0\rangle=\frac{1}{\sqrt{2}}|R\rangle+\frac{1}{\sqrt{2}}|B\rangle=\frac{1}{\sqrt{2}}(1,1)
\end{equation}
A generic state $p_v$ of both DM entity I and DM entity II is represented by the unit vector
\begin{equation} \label{v}
|v \rangle=\rho_R e^{i \theta_{R}}|R\rangle+\rho_B e^{i \theta_{B}}|B\rangle=(\rho_R e^{i \theta_{R}},\rho_B e^{i \theta_{B}})
\end{equation}
with $\rho_R,\rho_B\ge 0$, $\theta_R,\theta_B\in \Re$, $\rho_R^2+\rho_B^2=1$.

For every $i\in \{R,B\}$, the (subjective) probability $\mu_{v}(E_i)$ of drawing a ball of color $i$ in the state $p_v$ of either DM entity I or DM entity II is
\begin{equation}
\mu_{v}(E_i)=\langle v | P_{i} | v \rangle=|\langle i | v \rangle|^{2}=\rho_{i}^{2}
\end{equation}

Let us consider the representation of the acts $f_1$, $f_2$, $f_3$ and $f_4$ in Table 1, Sect. \ref{SEUT}. For given utility values $u(0)$ and $u(100)$, the acts $f_1$, $f_2$, $f_3$ and $f_4$ are respectively represented by the Hermitian operators
\begin{eqnarray}
\hat{F}_{1}&=&u(100)P_R+u(0)P_B \label{f1}\\
\hat{F}_{2}&=&u(100)P_R+u(0)P_B \label{f2}\\
\hat{F}_{3}&=&u(0)P_R+u(100)P_B\label{f3} \\
\hat{F}_{4}&=&u(0)P_R+u(100)P_B \label{f4}
\end{eqnarray}
The EU of $f_1,f_2,f_3$ and $f_4$ in a state $p_{v}$ of both DM entity I and DM entity II are respectively given by
\begin{eqnarray}
W_{v}(f_1)&=&\langle v| \hat{F}_{1}|v\rangle=\rho_{R}^{2}u(100)+\rho_{B}^{2}u(0)=\rho_{R}^{2}u(100)+(1-\rho_{R}^{2})u(0) \label{w1}\\
W_{v}(f_2)&=&\langle v| \hat{F}_{2}|v\rangle=\rho_{R}^{2}u(100)+(1-\rho_{R}^{2})u(0)  \label{w2} \\
W_{v}(f_3)&=&\langle v| \hat{F}_{3}|v\rangle=\rho_{R}^{2}u(0)+\rho_{B}^{2}u(100)=\rho_{R}^{2}u(0)+(1-\rho_{R}^{2})u(100) \label{w3}\\
W_{v}(f_4)&=&\langle v| \hat{F}_{4}|v\rangle=\rho_{R}^{2}u(0)+(1-\rho_{R}^{2})u(100) \label{w4}
\end{eqnarray}
where we have used (\ref{v}) and (\ref{f1})--(\ref{f4}).

Now, when a decision-maker is asked to ponder between the choice of acts $f_1$ and $f_2$, the pondering itself, before a decision is taken, defines a cognitive context, hence it may again change the state of DM entities I and II. Analogously, when a decision-maker is asked to ponder between the choice of acts $f_3$ and $f_4$, this defines a new cognitive context, before decision is taken, which may change the state of DM entities I and II. However, the pondering between $f_1$ and $f_2$ (and also the pondering between $f_3$ and $f_4$) will have different effects on DM entities I and II. Indeed, since act $f_1$ is ambiguous whereas $f_2$ is unambiguous, the comparison between $f_1$ and $f_2$ will determine a change of DM entity I from the state $p_{v_0}$ to a generally different state $p_{v_{12}}$, whereas the same comparison will leave DM entity II in the initial state $p_{v_0}$. Analogously, since $f_3$ is ambiguous whereas $f_4$ is unambiguous, the comparison between $f_3$ and $f_4$ will determine a change of DM entity I from the state $p_{v_0}$ to a generally different state $p_{v_{34}}$, whereas the same comparison will leave DM entity II in the initial state $p_{v_0}$.

Thus, the expected utilities in (\ref{w2}) and (\ref{w4}) in the state $p_0$ of DM entity II become $W_{v_0}(f_2)=W_{v_0}(f_4)=\frac{1}{2} (u(100)+u(0))$, which do not depend on the cognitive state of DM entity II, in agreement with the fact that $f_2$ and $f_4$ are unambiguous acts, while the EUs in (\ref{w1}) and (\ref{w3}) do depend on the final state of DM entity I, again in agreement with the fact that $f_1$ and $f_3$ are ambiguous acts.

Let us then prove that two ambiguity averse final states $p_{v_{12}}$ and $p_{v_{34}}$ of DM entity I exist such that the corresponding EUs satisfy the Ellsberg preferences in Sect. \ref{SEUT}, that is, $W_{v_{12}}(f_1)<W_{v_0}(f_2)$ and $W_{v_{34}}(f_3)<W_{v_0}(f_4)$. Indeed, suppose that the states $p_{v_{12}}$ and $p_{v_{34}}$ are represented, in the canonical ON basis of $\mathbb{C}^{2}$, by the unit vectors
\begin{eqnarray}
|v_{12}\rangle&=&(\sqrt{\alpha},\sqrt{1-\alpha}) \label{v12} \\
|v_{34}\rangle&=&(\sqrt{1-\alpha},-\sqrt{\alpha}) \label{v34} 
\end{eqnarray}
respectively, where $\alpha<\frac{1}{2}$. One preliminarily observes that the states $p_{v_{12}}$ and $p_{v_{34}}$ are represented by orthogonal vectors, that is, $\langle v_{12}| v_{34}\rangle=0$. Moreover, by using (\ref{w1})--(\ref{w4}), we have
\begin{eqnarray}
W_{v_{12}}(f_1)&=&\alpha u(100)+(1-\alpha)u(0)<\frac{1}{2}(u(100)+u(0))=W_{v_0}(f_2) \\
W_{v_{34}}(f_3)&=&(1-\alpha) u(0)+\alpha u(100)<\frac{1}{2}(u(100)+u(0))=W_{v_0}(f_4)
\end{eqnarray}
Hence, the ambiguity averse states $p_{v_{12}}$ and $p_{v_{34}}$ perfectly reproduce Ellsberg preferences against SEUT.

It remains to model the experimental data in Sect. \ref{SEUT}. To this end suppose that the {\it decision measurement between acts $f_1$ and $f_2$} is represented by the spectral family $\{ M, \mathbbmss{1}-M \}$, where the orthogonal projection operator $M$ projects onto the 1-dimensional subspace generated by the unit vector $|m\rangle=(\rho_m e^{i \theta_m}, \tau_m e^{i \phi_m})$ or, equivalently,
\begin{equation} \label{projM}
M=|m\rangle \langle m|=\left( 
\begin{array}{cc}
\rho_m^2 & \rho_m \tau_m e^{i (\theta_m-\phi_m)} \\
\rho_m \tau_m e^{-i (\theta_m-\phi_m)} & \tau_m^2 
\end{array} \right)
\end{equation}
Analogously, suppose that the {\it decision measurement between acts $f_3$ and $f_4$} is represented by the spectral family $\{ N, \mathbbmss{1}-N \}$, where the orthogonal projection operator $N$ projects onto the 1-dimensional subspace generated by the unit vector $|n\rangle=(\rho_n e^{i \theta_n}, \tau_n e^{i \phi_n})$ or, equivalently,
\begin{equation} \label{projN}
N=|n\rangle \langle n|=\left( 
\begin{array}{cc}
\rho_n^2 & \rho_n \tau_n e^{i (\theta_n-\phi_n)} \\
\rho_n \tau_n e^{-i (\theta_n-\phi_n)} & \tau_n^2 
\end{array} \right)
\end{equation}
It follows that the conditions
\begin{eqnarray}
\langle v_{12}|M|v_{12}\rangle&=&0.82 \label{exp12}\\
\langle v_{0}|M|v_{0}\rangle&=&0.50 \label{sym12}\\
\langle v_{12}|v_{12}\rangle&=&1 \label{norm12} \\
\langle v_{12}|N|v_{12}\rangle&=&0.84 \label{exp34} \\
\langle v_{0}|N|v_{0}\rangle&=&0.50 \label{sym34} \\
\langle v_{34}|v_{34}\rangle&=&1 \label{norm34}
\end{eqnarray}
must be satisfied by the parameters $\alpha<\frac{1}{2}$, $\rho_m, \tau_m, \rho_n, \tau_n \ge 0$, $\theta_m, \phi_m, \theta_n,\phi_n \in \Re$. Equations (\ref{exp12}) and (\ref{exp34}) are determined by empirical data, (\ref{norm12}) and (\ref{norm34}) are determined by normalization conditions, while (\ref{sym12}) and (\ref{sym34}) are determined by the fact that decision-makers who are not sensitive to ambiguity should overall be indifferent between $f_1$ and $f_2$, as well as between $f_3$ and $f_4$. Hence, on average, half respondents are expected to prefer $f_1$ ($f_3$) and the other half $f_2$ ($f_4$). To simplify the analysis, let us set $\theta_m=90^{\circ}$, $\theta_n=270^{\circ}$, $\phi_m=\phi_n=0$. Hence, we are left with a system of 6 equations in 5 unknown variables whose solution is
\begin{eqnarray}
\left\{ \begin{array}{lll}    
\alpha&=&0.14815 \\
\rho_m&=&0.21274 \\
\tau_m&=&0.97711 \\
\rho_n&=&0.99155 \\
\tau_n&=&0.12975
 \end{array} \right.
\end{eqnarray}
Hence, the orthogonal projection operators in (\ref{projM}) and (\ref{projN}) reproducing the experimental data in Sect. \ref{SEUT} are
\begin{eqnarray}
M&=&\left(
\begin{array}{cc}
0.04526 & 0.20787i\\
-0.20787i & 0.95474
\end{array} 
\right) \\
N&=&\left(
\begin{array}{cc}
0.98316 & -0.12865i\\
0.12865i & 0.01684
\end{array} \right)
\end{eqnarray}
This completes the construction of a quantum model for the Ellsberg two-urn experiment. As we can see, genuine quantum structures, namely, contextuality, superposition and intrinsically non-deterministic state change occur, while quantum probabilities crucially represent subjective probabilities.

\section{Modelling managerial and medical decisions}\label{conclusions}
The violation of SEUT in the Ellsberg two-urn experiment in Sect. \ref{SEUT} is paradigmatic, as the experiment provides the typical design one is confronted with in a variety of real life situations, like in economics, finance and even medicine, in which decision-makers have to choose between a risky prospect, i.e. a situation involving a known probability, and an ambiguous one, i.e. a situation involving an unknown ambiguity. DM tests in these domains confirm that people generally have an {\it ambiguity aversion attitude}, but they sometimes exhibit an {\it ambiguity seeking attitude}, also shifting from one attitude to the other depending on the degree of uncertainty involved in the decision. In this regard, let us consider the following medical situation example \cite{vc1999}.

Suppose your doctor tells you that there is a defined probability that you have a serious disease. So, you decide to consult other doctors: some of them believe that the probability is much greater while others believe that the probability is less. Which option would you ``prefer'', the former which is risky, or the latter which is ambiguous? Intuition suggests that the level of probability will play a crucial role in the final decision. Indeed, if the probability is low, it is reasonable to assume that a {\it fear effect} occurs in which you prefer the risky option to the ambiguous one, thus showing an ambiguity aversion behaviour. On the other side, if the probability is high, it is reasonable to assume that a {\it hope effect} instead occurs in which you prefer the ambiguous option, thus showing an ambiguity seeking behaviour.

Two experimental studies, \cite{vc1999} and \cite{hkk2002}, tested hope and fear effects in specific investment decisions involving business owners and managers. In managerial decisions, one typically measures the performance of an investment by comparing the value of a financial estimator, like the {\it internal rate of return} (IRR) or the {\it return on the investment} (ROI), with a targeted performance, or {\it benchmark}. If, say ROI, is above the benchmark, we say that a ``gain'' is realized, if ROI is instead below the benchmark, we say that a ``loss'' is realized. Thus, we may compare two options, a risky option, in which decision-makers know that the probability to have a gain is equal to $p$, with an ambiguous option, in which decision-makers only know that the probability to have a gain is between $p-\Delta$ and $p+\Delta$. Viscusi and Chesson \cite{vc1999} found that, if the probability of a gain is high, a fear effect occurs and people tend to be ambiguity averse. But, as this probability decreases, people become less ambiguity averse, reaching a {\it crossover point} in which they become ambiguity seeking, which suggests a shift from a fear to a hope effect. Viceversa, if the probability of a loss is high, a hope effect occurs and people tend to be ambiguity seeking. But, as this probability decreases, people become less ambiguity seeking, again reaching a crossover point in which they become ambiguity averse, which suggests a shift from a hope to a fear effect \cite{vc1999}. At high probabilities, this behaviour was confirmed by the test of Ho, Keller and Keltyka \cite{hkk2002}.

We note that the quantum-theoretic framework of Sect. \ref{QEUT} can be naturally applied to both ambiguity averse and ambiguity seeking attitudes, as the type of state change of the DM entity induced by the interaction with the cognitive context in the DM process determines individual attitudes towards ambiguity. In addition, the Hilbert space model of Sect. \ref{quantumellsberg} can be applied to represent the empirical data collected in both tests in \cite{vc1999} and \cite{hkk2002}. These tests are indeed variants of the two-urn experiment in which each time a risky option is compared with an ambiguous one, hence one can construct Hilbert space states, operators and quantum probabilities that faithfully represent data sets.

The applicability of the theoretic framework of Sect. \ref{QEUT} to a wide range of DM situations was a priori expected. Indeed, the quantum-theoretic framework and the quantum models derived from it
are not ad hoc models, in the sense that they were not constructed on purpose to accommodate the Ellsberg paradox and fit the corresponding experiments. They were instead generated from the investigation of quantum theory as a unified and general theory to model human DM under uncertainty, hence they can be applied in principle to any type of empirical situation where people take decisions in the presence of uncertainty, like investment and medical situations, which is what we plan to do in future research.

\section*{Acknowledgments}
The author is greatly indebted with  Prof. Diederik Aerts and Dr. Massimiliano Sassoli de Bianchi for reading the manuscript and providing a number of valuable comments and suggestions for improvement.

\end{document}